\documentclass[11pt,a4paper]{article}
\usepackage[margin=25mm]{geometry}
\usepackage{fontspec}
\newfontfamily\bengalifont[Path=fonts/,Script=Bengali,Scale=1.03]{NotoSerifBengali-Regular.ttf}
\newfontfamily\ipafont[Path=fonts/,Scale=0.92]{DejaVuSans.ttf}
\newcommand{\bn}[1]{{\bengalifont #1}}
\newcommand{\ipa}[1]{{\ipafont #1}}
\usepackage{graphicx}
\usepackage{xcolor}
\usepackage{booktabs,array,tabularx}
\usepackage{caption,subcaption}
\usepackage{listings}
\usepackage{placeins}
\usepackage[numbers,sort&compress]{natbib}
\usepackage{hyperref}
\definecolor{linkblue}{HTML}{204D72}
\usepackage{wrapfig}
\usepackage{needspace}
\definecolor{notefg}{HTML}{76500D}
\hypersetup{colorlinks=true,linkcolor=linkblue,citecolor=linkblue,urlcolor=linkblue,
  pdftitle={PromitoLipi: A versatile offline dataset of handwritten Bangla words and paragraphs},
  pdfauthor={Faias Satter and Sk. Md. Masudul Ahsan}}
\renewcommand{\arraystretch}{1.10}
\newcolumntype{Y}{>{\raggedright\arraybackslash}X}

\title{\LARGE\bfseries PromitoLipi: A versatile offline dataset of handwritten Bangla words and paragraphs}
\author{
  Faias Satter\textsuperscript{*}
  \quad Noor Masrur
  \quad Sk. Md. Masudul Ahsan\\[4pt]
  \small Department of Computer Science and Engineering\\
  \small Khulna University of Engineering \& Technology\\
  \small Khulna-9203, Bangladesh\\[4pt]
  \small
  \href{mailto:faiaspromit@gmail.com}{faiaspromit@gmail.com}
  \quad
  \href{mailto:noor.masrur.bd@gmail.com}{noor.masrur.bd@gmail.com}
  \quad
  \href{mailto:smahsan@cse.kuet.ac.bd}{smahsan@cse.kuet.ac.bd}\\[3pt]
  \small \textsuperscript{*}Corresponding author
}

\hypersetup{
  pdfauthor={Faias Satter, Noor Masrur, and Sk. Md. Masudul Ahsan}
}
\date{}
\begin{document}
\maketitle
\begin{abstract}
PromitoLipi is a publicly available offline dataset of handwritten Bangla, organized around two stages of optical character recognition: locating words on a page and recognizing the characters within them. PromitoLipi1.1 contains 80 single-page
handwritten paragraphs with 7,231 words. To our knowledge,
it is the first public Bangla handwriting dataset in which
both paper and ink colors were deliberately varied.
Pages were written using different pens and pencils and
captured with scanners and smartphone cameras, preserving
shadows, background elements, and writing corrections.
These variations support research on binarization and
word segmentation. PromitoLipi2.1 contains 9,830 binarized
images of words, non-word character combinations, numerals,
and isolated characters, with 24,050 manually annotated
bounding boxes across 88 visual classes, including 17
selected conjuncts. Its non-word examples support recognition
evaluation beyond a fixed vocabulary of valid Bangla words.
The author-collected handwriting in both datasets comes from
53 writers; it makes up approximately 70\% of PromitoLipi2.1,
and the remainder was derived from CMATERdb. This article
describes the collection, preparation, and annotation
procedures, summarizes results from earlier studies, and
discusses reuse limitations. The data are available through
Mendeley Data, Version 3, at
\href{https://doi.org/10.17632/fnw59h7y89.3}
{doi:10.17632/fnw59h7y89.3}.
\end{abstract}


\section{Introduction and value of the data}
\label{sec:introduction}

Handwritten Bangla document analysis requires locating text
within a page and recognizing characters whose shapes, sizes,
spacing, and joining patterns vary among writers. Differences
in paper, ink, illumination, and image acquisition further
affect the appearance of the writing. Data that capture these
variations are useful for studying both text segmentation
and recognition.

This article presents PromitoLipi~\cite{promitolipi2026},
a publicly available resource organized around these two
stages of Bangla OCR. PromitoLipi1.1 preserves handwriting
in its page context for studying line and word segmentation.
PromitoLipi2.1 provides cropped handwriting images
with character-component bounding-box annotations
for studying character detection and word recognition.

To our knowledge, PromitoLipi1.1 is the first publicly available Bangla handwriting dataset in which both paper color and ink color were deliberately varied. This produces a wide range of contrast between text and background, which is useful for studying binarization and segmentation. The collection descriptions of BanglaWriting~\cite{mridha2021banglawriting}, BN-HTRd~\cite{rahman2022bnhtrd}, and the Boise State dataset~\cite{majid2018boise} do not report comparable variation.

PromitoLipi2.1 includes non-word character combinations on purpose, so its recognition targets are not limited to Bangla words. In comparison, BanglaWriting contains text written on freely chosen topics~\cite{mridha2021banglawriting}, the text in BN-HTRd comes from news articles~\cite{rahman2022bnhtrd}, and the essay pages of the Boise State dataset use a fixed passage~\cite{majid2018boise}. Because these combinations are not words, a recognizer cannot guess them from familiar vocabulary and must read each character as written.

\begin{wrapfigure}{r}{0.50\textwidth}
    \centering
    \includegraphics[width=\linewidth]
        {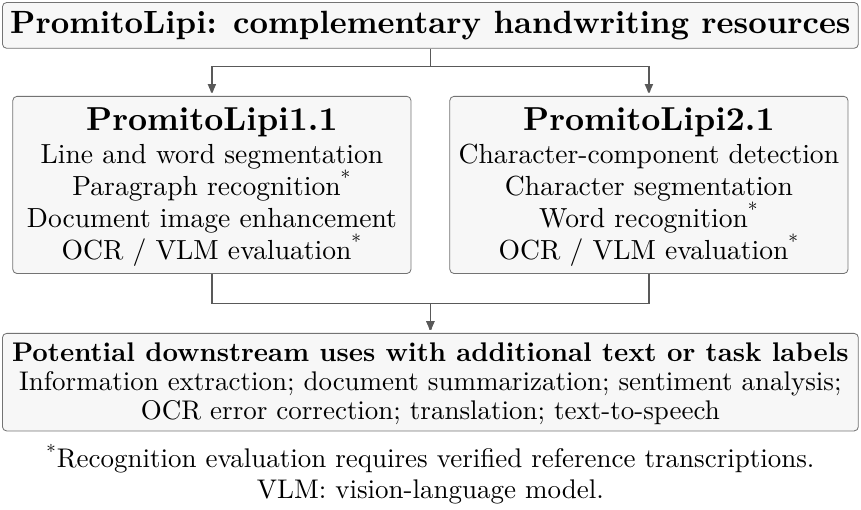}
    \caption{Potential applications of PromitoLipi1.1 and
PromitoLipi2.1. Downstream language tasks require additional
labels or processing.}
    \label{fig:applications}
\end{wrapfigure}

PromitoLipi was developed as part of an undergraduate project
on Bangla OCR at Khulna University of Engineering \& Technology
in 2022--2023, documented in the first author's
thesis~\cite{satter2023thesis}. The project subsequently led
to three related publications addressing different stages
of the OCR pipeline: \citet{satter2023segmentation} presented
a word segmentation method for pages with varying paper
and ink colors; \citet{satter2023recognition} investigated
character-component detection and word recognition; and
\citet{masrur2023dissimilarity} developed an approach to
correcting OCR errors. The present article documents
the datasets developed through this project, including
their collection, preparation, annotation, and reuse. Figure~\ref{fig:applications} summarizes potential
applications of PromitoLipi.

\begin{wrapfigure}{r}{0.48\textwidth}
\vspace{-22pt}
\centering
\includegraphics[width=\linewidth]{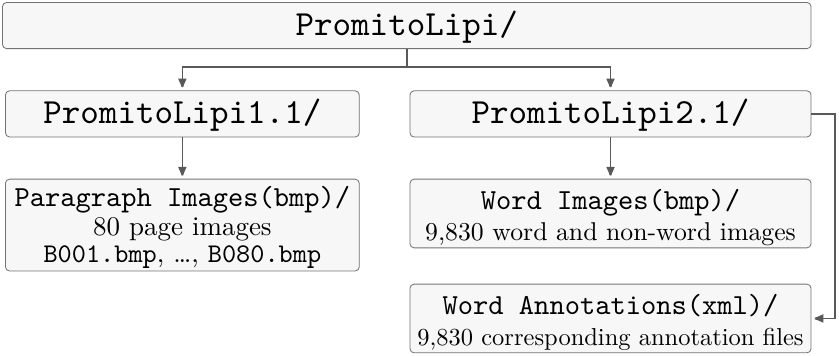}
\caption{Directory structure of PromitoLipi, showing paragraph
images in PromitoLipi1.1 and binarized handwriting images
with matching XML annotation files in PromitoLipi2.1.}
\vspace{-3pt}
\label{fig:directory}
\end{wrapfigure}

\section{Dataset overview and access}
\label{sec:overview}
This article describes Version 3 of PromitoLipi on Mendeley
Data~\cite{promitolipi2026}, available at
\url{https://data.mendeley.com/datasets/fnw59h7y89/3} and
comprising PromitoLipi1.1 and PromitoLipi2.1.
Approximately 70\% of PromitoLipi2.1 was prepared from
author-collected handwriting, comprising pages included
in PromitoLipi1.1 and separately collected word-list pages.
The remaining approximately 30\% was derived from selected
CMATERdb collections~\cite{cmaterdb,das2012digits,das2014compound}.
PromitoLipi1.1 and PromitoLipi2.1 therefore share part of
their source handwriting. Table~\ref{tab:overview}
summarizes the two datasets. Figure~\ref{fig:directory} shows the core file organization. PromitoLipi2.1 images and their annotations share the same filename stem, for example \texttt{7435.bmp} and \texttt{7435.xml}.

\begin{table}[tbp]
\centering\small
\caption{Overview of the released datasets.}
\label{tab:overview}
\setlength{\tabcolsep}{1pt}
\begin{tabularx}{\linewidth}{@{}p{2.5cm}YY@{}}
\toprule
Property & PromitoLipi1.1 & PromitoLipi2.1\\\midrule
Content & Single-page paragraphs & Words, non-word character combinations, numerals, and isolated characters\\
Images & 80 & 9,830\\
Reported units & 7,231 words & 24,050 labeled instances\\
Image format & BMP & BMP\\
Processing & Cropped pages; some shadows and backgrounds retained & Segmented, refined, and binarized\\
Annotations & Per-image word count and paragraph type & Matching VOC-style XML files with class IDs and bounding boxes\\
Visual classes & Not applicable & 88\\
Sources & Collected handwriting & 70\% collected; 30\% CMATERdb-derived\\\bottomrule
\end{tabularx}
\end{table}

\section{Data description}

\begin{figure}[htbp]
\centering
\begin{subfigure}[t]{0.24\linewidth}
\centering
\includegraphics[width=\linewidth]{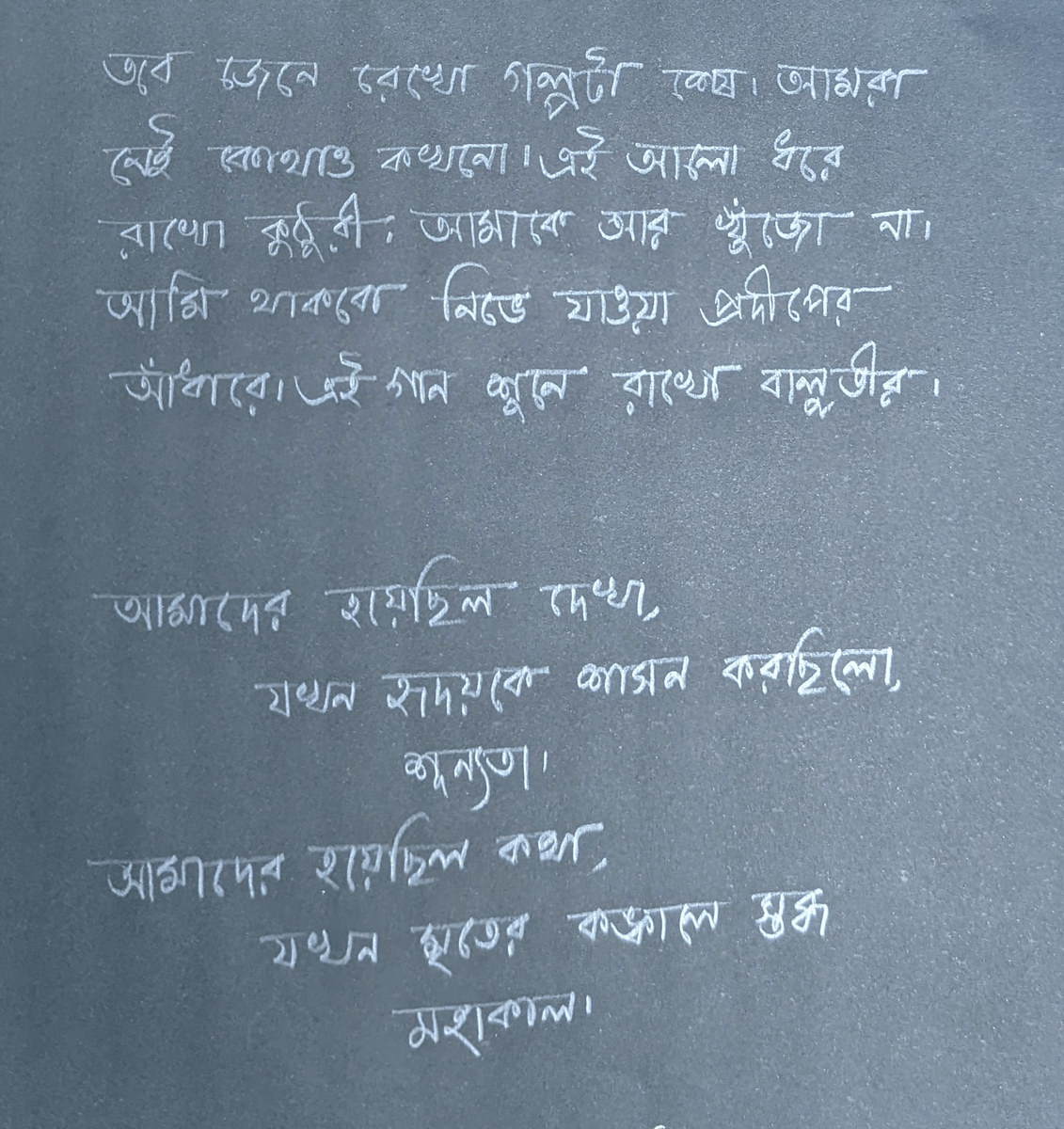}
\caption{B008.bmp; 45 words}
\end{subfigure}\hfill
\begin{subfigure}[t]{0.24\linewidth}
\centering
\includegraphics[width=\linewidth]{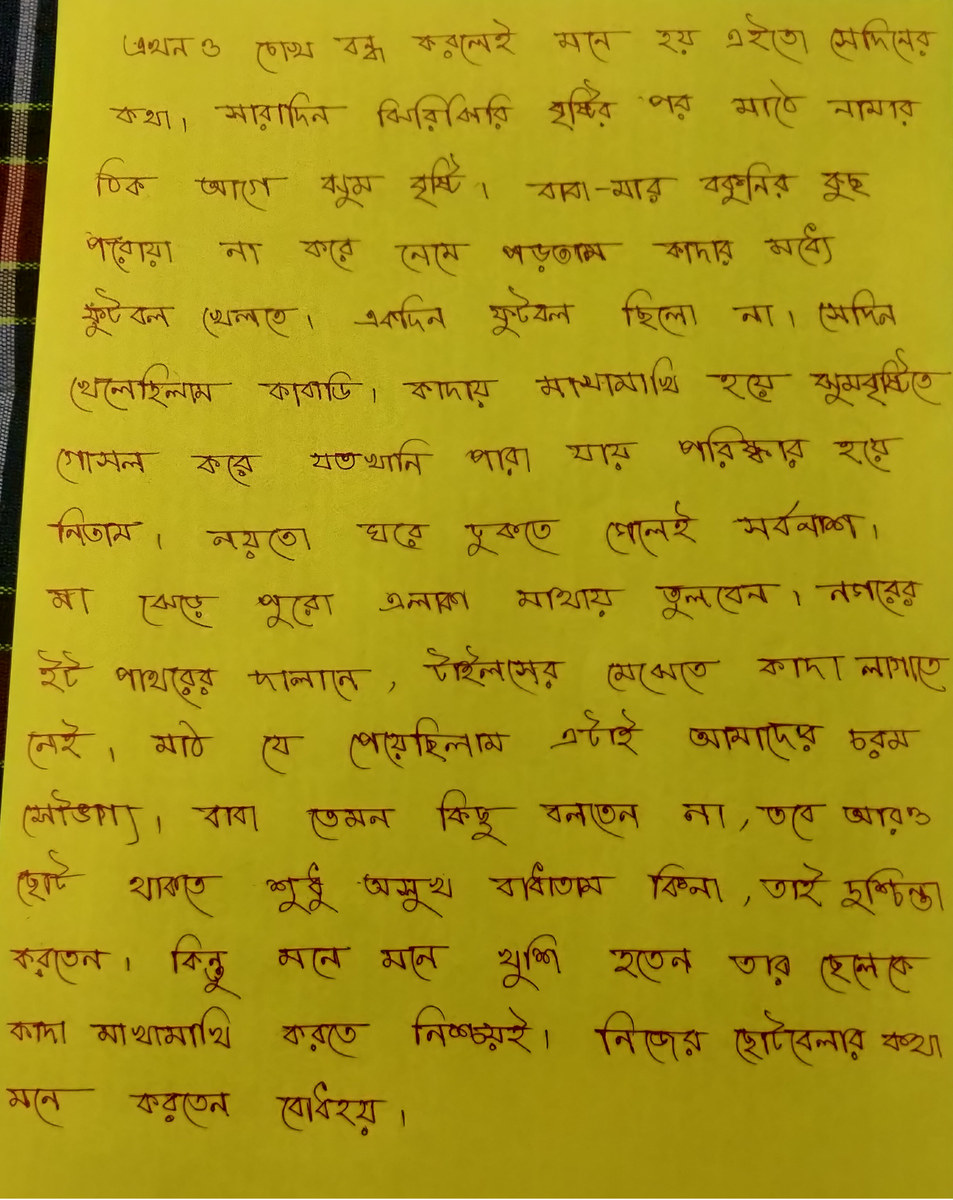}
\caption{B072.bmp; 111 words}
\end{subfigure}\hfill
\begin{subfigure}[t]{0.24\linewidth}
\centering
\includegraphics[width=\linewidth]{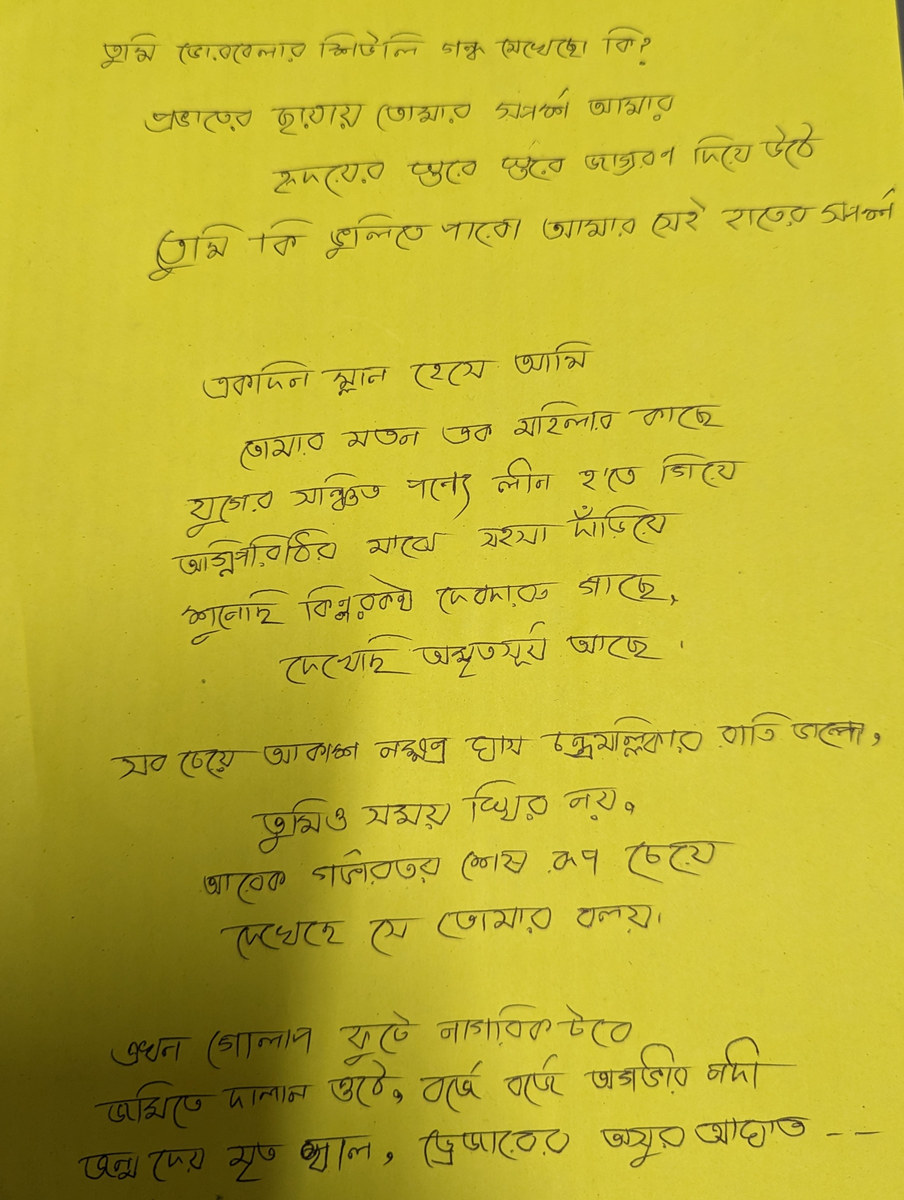}
\caption{B077.bmp; 93 words}
\end{subfigure}\hfill
\begin{subfigure}[t]{0.24\linewidth}
\centering
\includegraphics[width=\linewidth]{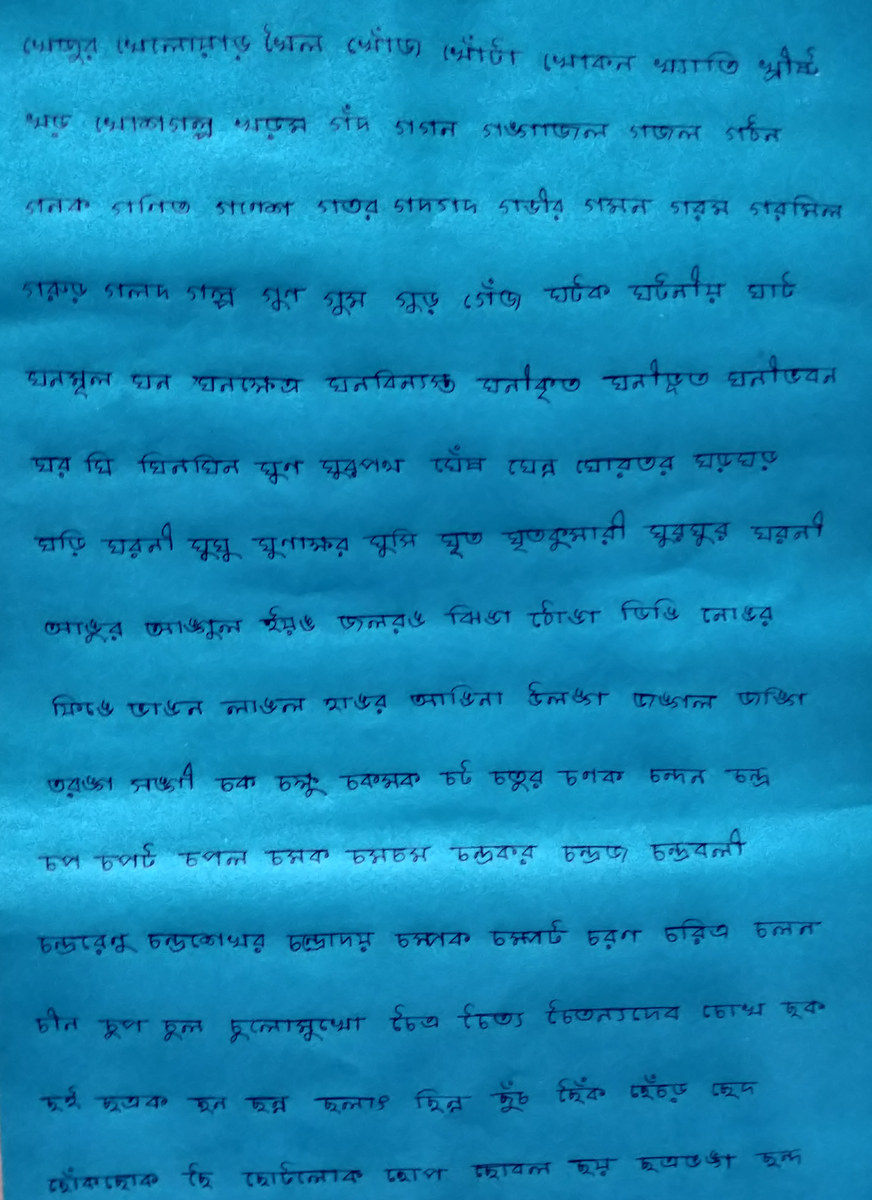}
\caption{B080.bmp; 129 words}
\end{subfigure}
\caption{Paragraph examples from PromitoLipi1.1 with filenames and word counts.}
\label{fig:paragraphs}
\end{figure}

\begin{figure}[p]
\centering
\begin{minipage}[b]{.275\linewidth}
\vspace{0pt}\centering
\includegraphics[width=\linewidth]{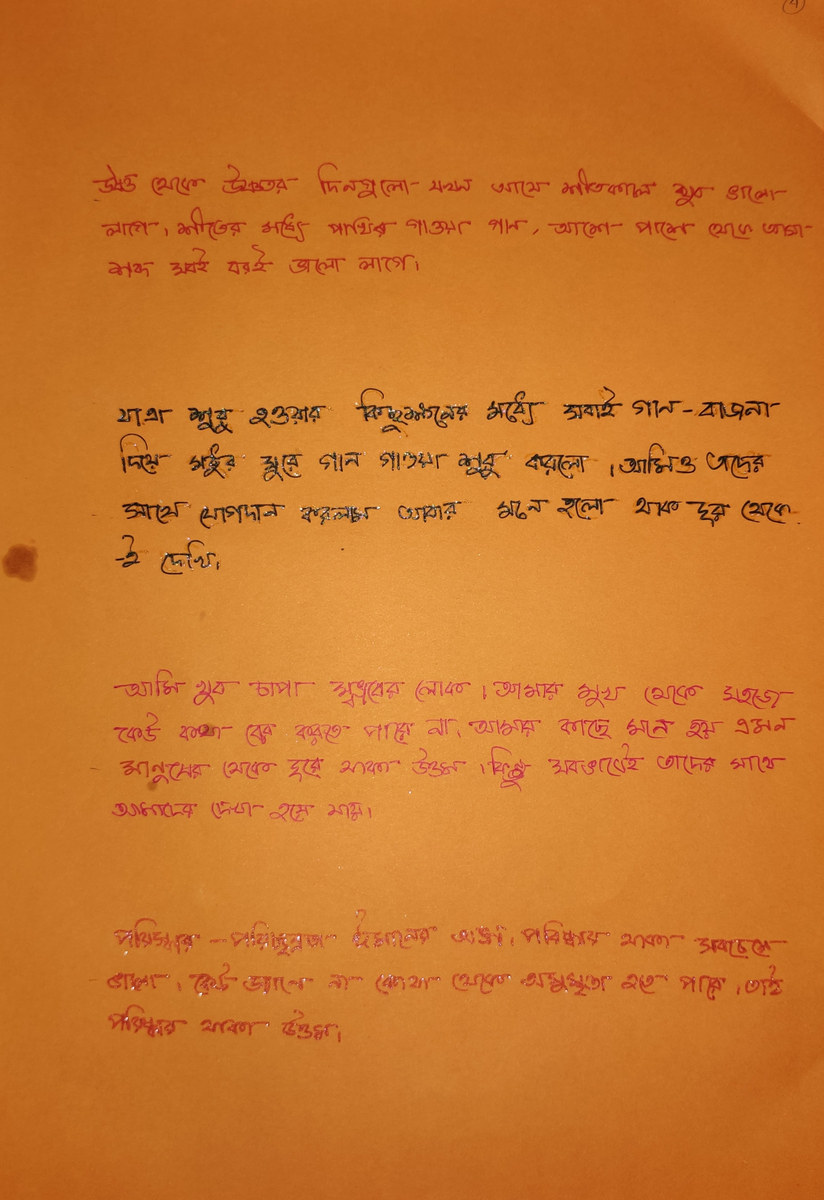}
\par\smallskip\small (a) B019.bmp; 103 words
\end{minipage}\hfill
\begin{minipage}[b]{.335\linewidth}
\vspace{0pt}\small\raggedright
\bn{উষ্ণ থেকে উষ্ণতর দিনগুলো যখন আসে শীতকালে খুব ভালো লাগে। শীতের মধ্যে পাখির গাওয়া গান, আশে পাশে থেকে আসা শব্দ সবই বরই ভালো লাগে।}

\bn{যাত্রা শুরু হওয়ার কিছু ক্ষণের মধ্যে সবাই গান-বাজনা দিয়ে মধুর গান গাওয়া শুরু করলো। আমিও তাদের সাথে যোগদান করলাম আবার মনে হলো থাক দূর থেকেই দেখি।}

\bn{আমি খুব চাপা স্বভাবের লোক। আমার মুখ থেকে সহজে কেউ কথা বের করতে পারে না। আমার কাছে মনে হয় এমক্ন মানুষের থেকে দূরে থাকা উত্তম। কিন্তু সবভাবেই তাদের সাথে আমাদের দেখা হয়ে যায়।}

\bn{পরিষ্কার-পরিচ্ছিন্নতা ইমানের অঙ্গ। পরিষ্কার থাকা সবচেয়ে ভালো। কেউ জানে না কোথা থেকে অসুস্থতা হতে পারে। তাই পরিষ্কার থাকা উত্তম।}
\par\smallskip\normalfont\small (b) Bangla transcription
\end{minipage}\hfill
\begin{minipage}[b]{.335\linewidth}
\vspace{0pt}\small\raggedright
\ipa{/uʃno t̪ʰeke uʃnot̪ɔr d̪ingulo ɟɔkʰon aʃe ʃit̪kale kʰub bʱalo lage\bn{।} ʃit̪er mod̪d̪ʱe pakʰir ga͡o̯ʷa gan, aʃe paʃe t̪ʰeke aʃa ʃɔbd̪o ʃɔbi bɔro͡i̯ bʱalo lage\bn{।}}

\ipa{ɟat̪ra ʃuru hoʷar kicʰukkʰoner mod̪d̪ʱe ʃɔba͡i̯ gan-baɟna d̪iʲe mod̪ʱur gan ga͡o̯ʷa ʃuru korlo\bn{।} ami͡o̯ t̪ad̪er ʃat̪ʰe ɟogd̪an korlam abar mone holo t̪ʰak d̪ur t̪ʰeke͡i̯ d̪ekʰi\bn{।}}

\ipa{ami kʰub capa ʃɔbʱaber lok\bn{।} amar mukʰ t̪ʰeke ʃɔhoɟe ke͡u̯ kɔt̪ʰa ber kort̪e pare na\bn{।} amar kacʰe mone hɔʲ emokko manuʃer t̪ʰeke d̪ure t̪ʰaka ut̪t̪om\bn{।} kint̪u ʃɔbbʱabe͡i̯ t̪ad̪er ʃat̪ʰe amad̪er d̪ækʰa hoʲe ɟaʲ\bn{।}}

\ipa{poriʃkar-poriccʰinnot̪a imaner ɔŋgo\bn{।} poriʃkar t̪ʰaka ʃɔbceʲe bʱalo\bn{।} ke͡u̯ ɟane na kot̪ʰa t̪ʰeke ɔʃust̪ʰot̪a hot̪e pare\bn{।} t̪a͡i̯ poriʃkar t̪ʰaka ut̪t̪om\bn{।}/}
\par\smallskip\normalfont\small (c) IPA transcription
\end{minipage}
\caption{Sample \texttt{B019.bmp} from PromitoLipi1.1 with Bangla and corresponding IPA transcriptions.}
\label{fig:transcription}
\end{figure}

\subsection{PromitoLipi1.1}
PromitoLipi1.1 contains 80 paragraph images with 7,231 words. Each page contains one individual's handwriting. Figure~\ref{fig:paragraphs} shows examples with different paper and ink colors. The complete filename and word-count inventory is provided in Appendix~\ref{app:paragraphs}. Paragraphs are labeled as coherent (C) when their sentences
are meaningfully related to one another, or incoherent (I)
when they consist of unrelated words. Figure~\ref{fig:transcription} shows one sample with a Bangla
transcription and an IPA transcription to illustrate its
pronunciation for readers unfamiliar with the
script~\cite{zubaer2020ipa}. These transcriptions are provided
for this example, not for every page in the dataset.

\begin{wrapfigure}{r}{0.62\textwidth}
\vspace{-2pt}
\centering
\setlength{\tabcolsep}{2pt}
\begin{tabular}{@{}ccc@{}}
\includegraphics[width=.30\linewidth,height=2.15cm,keepaspectratio]{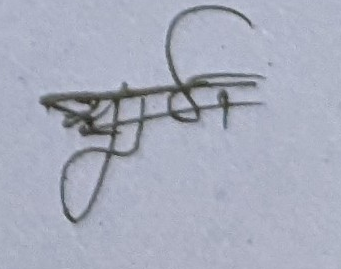}&
\includegraphics[width=.30\linewidth,height=2.15cm,keepaspectratio]{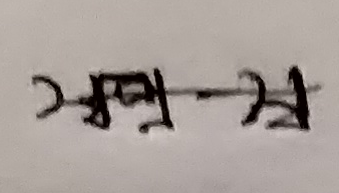}&
\includegraphics[width=.30\linewidth,height=2.15cm,keepaspectratio]{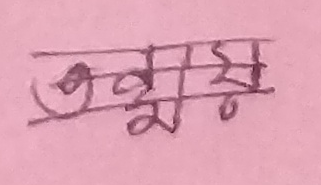}\\
\footnotesize (a) Crossed out &
\footnotesize (b) Crossed out &
\footnotesize (c) Crossed out\\[7pt]
\includegraphics[width=.30\linewidth,height=1.7cm,keepaspectratio]{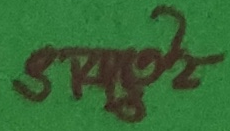}&
\includegraphics[width=.30\linewidth,height=1.7cm,keepaspectratio]{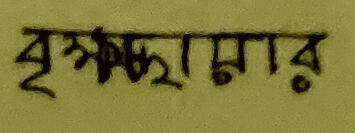}&
\includegraphics[width=.30\linewidth,height=1.7cm,keepaspectratio]{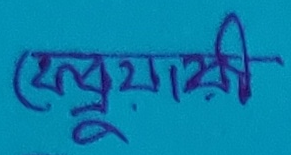}\\
\footnotesize (d) Overwritten &
\footnotesize (e) Overwritten &
\footnotesize (f) Overwritten
\end{tabular}
\caption{Selected writing corrections in PromitoLipi1.1:
cross-outs from (a) \texttt{B003.bmp},
(b) \texttt{B013.bmp}, and (c) \texttt{B063.bmp};
overwriting from (d) \texttt{B010.bmp},
(e) \texttt{B057.bmp}, and (f) \texttt{B039.bmp}.}
\vspace{-25pt}
\label{fig:artifacts}
\end{wrapfigure}

\paragraph{Writing corrections and image artifacts.}

PromitoLipi1.1 retains writing corrections, including
crossed-out and overwritten text. For example,
\texttt{B001.bmp} and \texttt{B028.bmp} contain four
and two cross-out occurrences, respectively, while
\texttt{B036.bmp} and \texttt{B060.bmp} contain eight
and three instances of overwriting, respectively.
Figure~\ref{fig:artifacts} presents additional examples,
with their source filenames identified in the caption.

\subsection{PromitoLipi2.1}
\begin{figure}[htbp]
\centering\footnotesize
\setlength{\tabcolsep}{2pt}
\renewcommand{\tabularxcolumn}[1]{m{#1}}
\begin{tabularx}{\linewidth}{@{}
    >{\centering\arraybackslash}m{1.7cm}
    >{\centering\arraybackslash}m{1.9cm}
    Y
    >{\centering\arraybackslash}m{1.7cm}
    >{\centering\arraybackslash}m{1.9cm}
    Y@{}}
\toprule
Image & Annotation overlay & Description &
Image & Annotation overlay & Description\\
\midrule

\includegraphics[width=1.7cm,height=1.05cm,keepaspectratio]{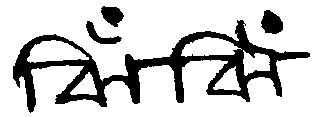} &
\includegraphics[width=1.9cm,height=1.05cm,keepaspectratio]{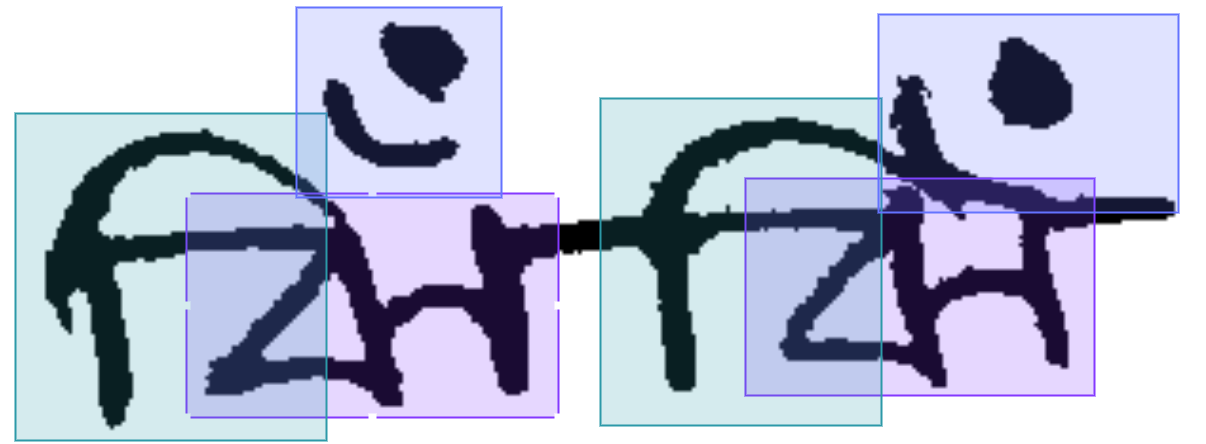} &
\texttt{20236.bmp}\newline
Classes: \bn{ি, ঁ, ঝ, ি, ঁ, ঝ}\newline
IDs: 61, 49, 19, 61, 49, 19\newline
IPA: \ipa{/i/, / ̃/, /ɟʱ/, /i/, / ̃/, /ɟʱ/}\newline
Lexical word &
\includegraphics[width=1.7cm,height=1.05cm,keepaspectratio]{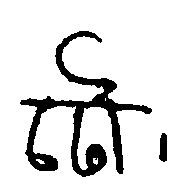} &
\includegraphics[width=1.9cm,height=1.05cm,keepaspectratio]{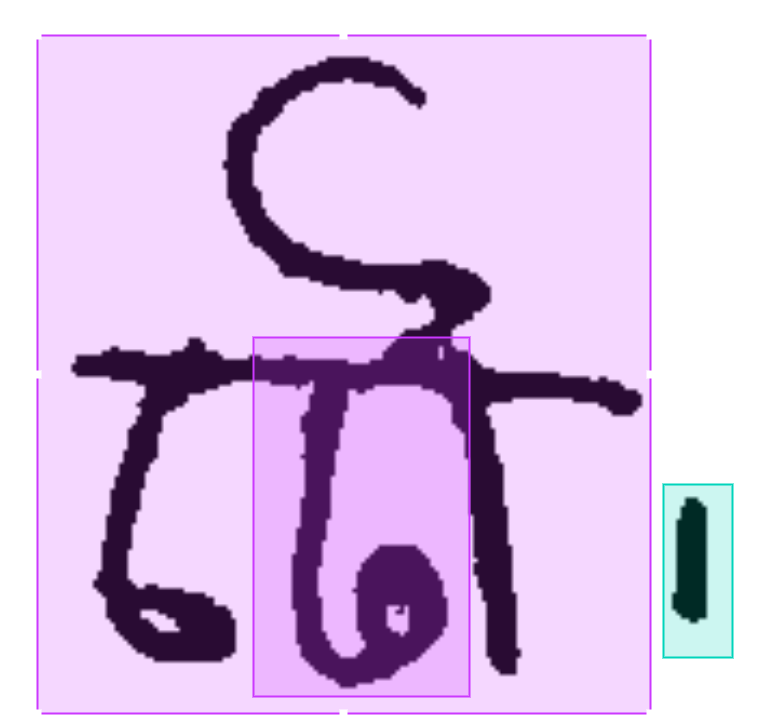} &
\texttt{21429.bmp}\newline
Classes: \bn{ৌ, ঢ, ।}\newline
IDs: 91, 24, 92\newline
IPA: \ipa{/o͡u̯/, /dʱo/, /\bn{।}/}\newline
Non-word\\[7pt]

\includegraphics[width=1.7cm,height=1.05cm,keepaspectratio]{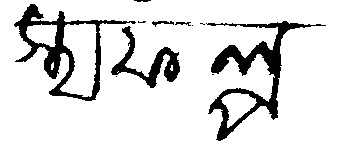} &
\includegraphics[width=1.9cm,height=1.05cm,keepaspectratio]{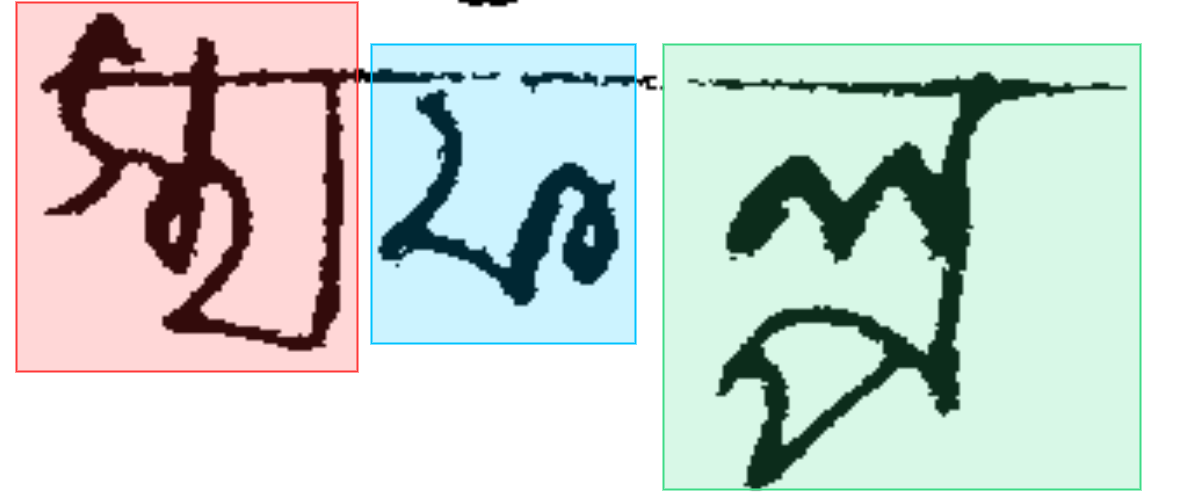} &
\texttt{21986.bmp}\newline
Classes: \bn{স্থ, ফ, ল্প}\newline
IDs: 72, 32, 84\newline
IPA: \ipa{/st̪ʰɔ/, /pʰo/, /lpo/}\newline
Non-word &
\includegraphics[width=1.7cm,height=1.05cm,keepaspectratio]{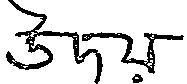} &
\includegraphics[width=1.9cm,height=1.05cm,keepaspectratio]{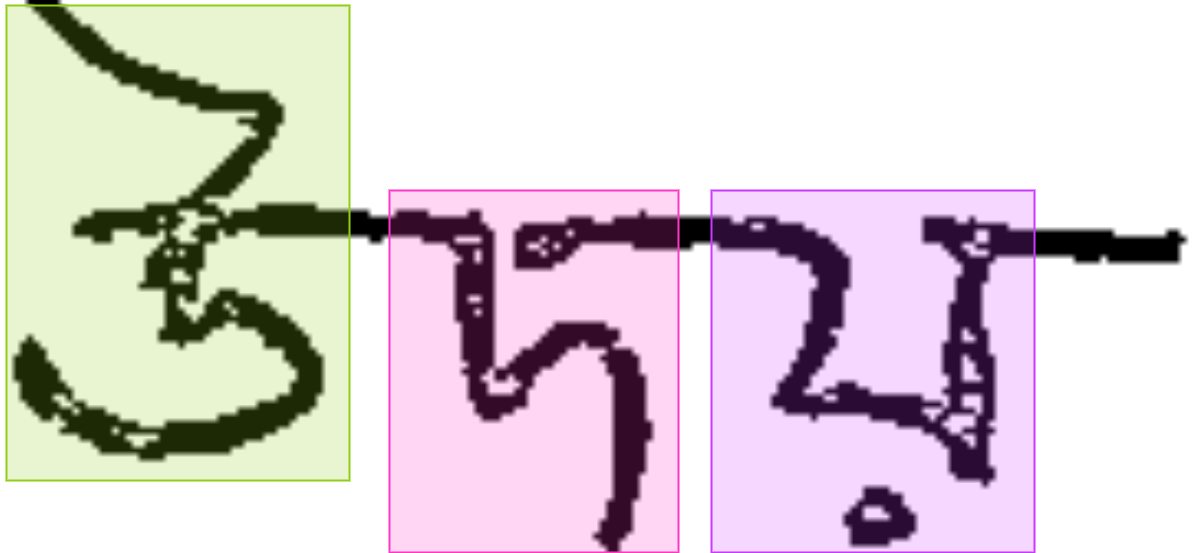} &
\texttt{4453.bmp}\newline
Classes: \bn{উ, দ, য়}\newline
IDs: 4, 28, 45\newline
IPA: \ipa{/u/, /d̪ɔ/, /ʲ/}\newline
Lexical word\\[7pt]

\includegraphics[width=1.7cm,height=1.05cm,keepaspectratio]{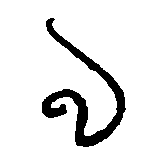} &
\includegraphics[width=1.9cm,height=1.05cm,keepaspectratio]{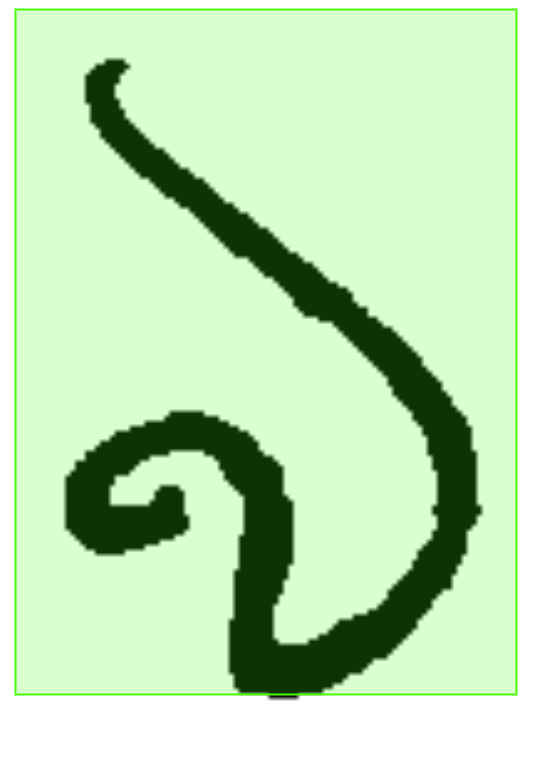} &
\texttt{10800.bmp}\newline
Classes: \bn{৯}\newline
IDs: 59\newline
IPA: \ipa{/9/}\newline
Single numeral &
\includegraphics[width=1.7cm,height=1.05cm,keepaspectratio]{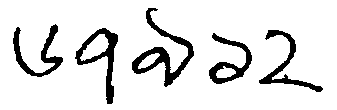} &
\includegraphics[width=1.9cm,height=1.05cm,keepaspectratio]{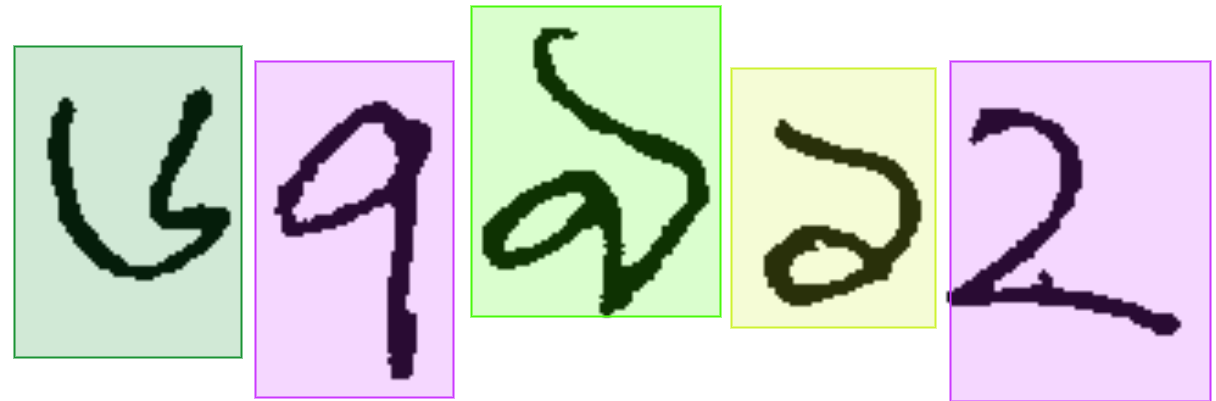} &
\texttt{20774.bmp}\newline
Classes: \bn{৬, ৭, ৯, ১, ২}\newline
IDs: 56, 57, 59, 51, 52\newline
IPA: \ipa{/6/, /7/, /9/, /1/, /2/}\newline
Numeral sequence\\[7pt]

\includegraphics[width=1.7cm,height=1.05cm,keepaspectratio]{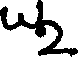} &
\includegraphics[width=1.9cm,height=1.05cm,keepaspectratio]{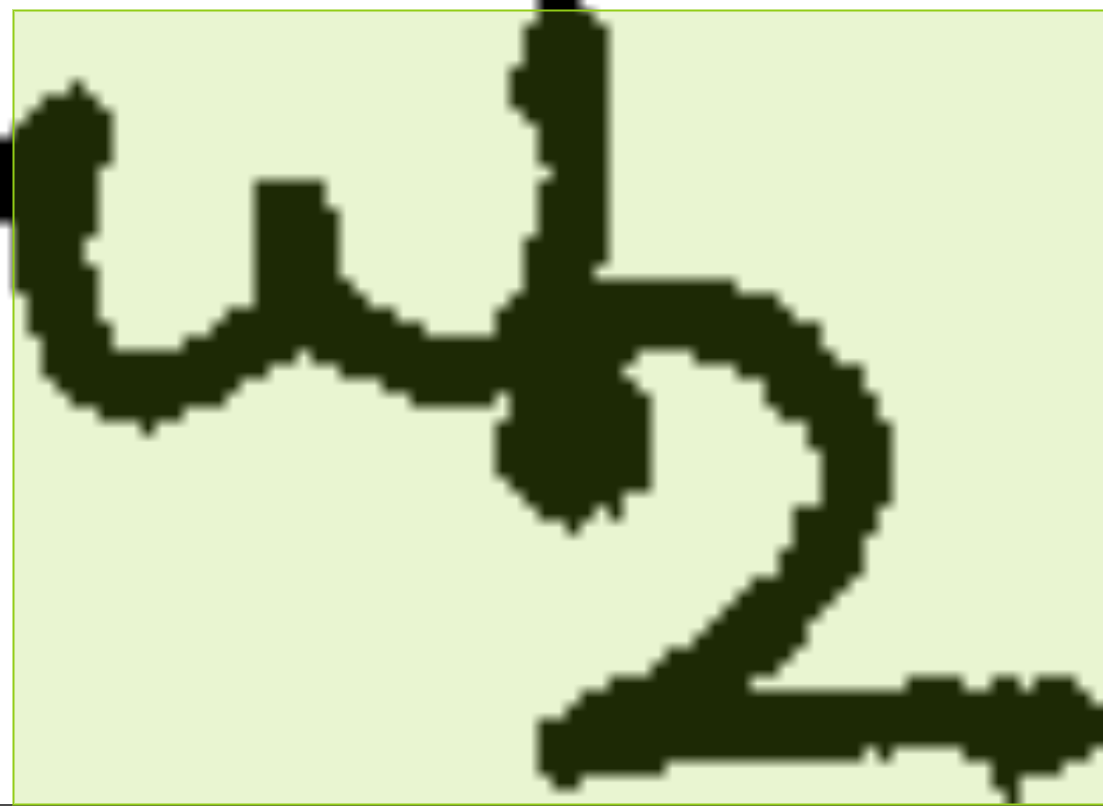} &
\texttt{B20265.bmp}\newline
Classes: \bn{স্থ}\newline
IDs: 72\newline
IPA: \ipa{/st̪ʰɔ/}\newline
Single conjunct &
\includegraphics[width=1.7cm,height=1.05cm,keepaspectratio]{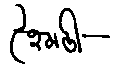} &
\includegraphics[width=1.9cm,height=1.05cm,keepaspectratio]{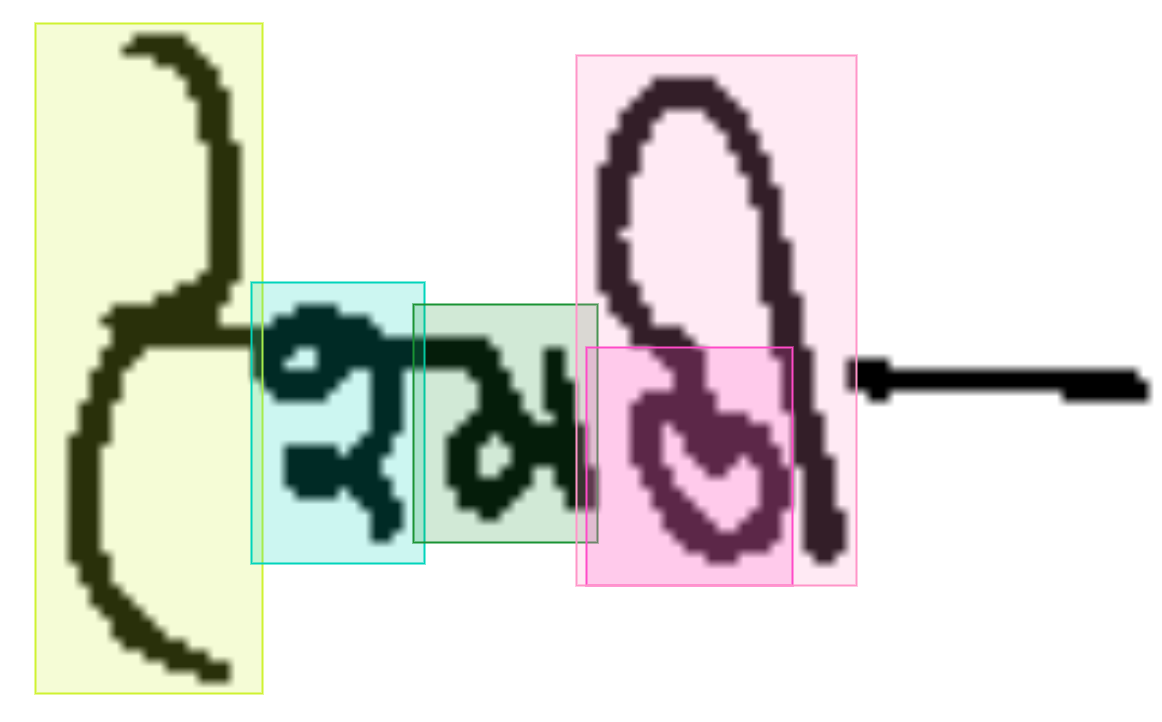} &
\texttt{B1366.bmp}\newline
Classes: \bn{ৈ, হ, ম, ন্ত, ী}\newline
IDs: 89, 42, 35, 69, 62\newline
IPA: \ipa{/o͡i̯/, /ho/, /mɔ/, /nt̪/, /i/}\newline
Lexical word\\[7pt]

\bottomrule
\end{tabularx}
\caption{Examples from PromitoLipi2.1 showing handwriting images,
component annotations, and corresponding class labels.
Colors distinguish the displayed bounding boxes.}
\label{fig:words}
\end{figure}

PromitoLipi2.1 contains 9,830 images of handwritten Bangla
words, non-word character combinations, numerals, numeral
sequences, and isolated characters, including conjuncts. Figure~\ref{fig:words} presents eight examples from
PromitoLipi2.1 with their corresponding annotations.
The dataset contains 24,050 annotated instances across
88 classes, covering vowels, consonants, diacritics,
numerals, punctuation, and 17 selected conjuncts.
Each instance is marked by a rectangular bounding box
and assigned a class label.

Class counts were not equalized during collection.
The collection includes many everyday Bangla words,
and vowel signs such as \bn{া}, \bn{ে}, and \bn{ি}
occur frequently in the collected handwriting.
The most frequent class, \bn{া}, has 2,040 annotated
instances, compared with 99 for the least frequent
class, \bn{ৃ}. Users may apply class weighting or
resampling during training. The complete class inventory
and occurrence counts are provided in
Appendix~\ref{app:classes}.

\section{Collection protocol}
\subsection{Participants and sampling}
A total of 53 writers contributed handwriting to PromitoLipi1.1
and the author-collected portion of PromitoLipi2.1.
Writers were recruited through the authors' personal and
academic networks, in person or through social media, using
convenience sampling. Contributors included school students,
older adults, and university students from
Khulna University of Engineering \& Technology
(KUET), North South University (NSU), Bangladesh University of
Business and Technology (BUBT), Brac University, Shahjalal
University of Science and Technology (SUST), and Military
Institute of Science and Technology (MIST). These institutions
were included because the authors could reach writers there
directly, rather than to represent the wider population.
Recruitment through academic networks explains the emphasis
on university students.

The approximate age distribution was 10--14 years (3 writers),
15--19 years (4), 20--30 years (34), 40--45 years (5), and
55--60 years (7).
Writers aged 20--30 formed the majority; the collection therefore
has limited coverage of other age groups, and its emphasis on
university students leaves other educational backgrounds
underrepresented.

\subsection{Writing procedure and image acquisition}
For PromitoLipi1.1, writers used A4-sized paper in white,
black, blue, red, yellow, and orange. Writing instruments
included gel, glitter, and ballpoint pens, as well as pencils.
Pen ink colors included black, blue, orange, green, red,
and pink. These materials were deliberately varied to
introduce differences in stroke width, texture, and
text--background contrast. Pencil strokes can be light
and grainy, while glitter ink can produce uneven reflections
in photographs.

An author was present while many participants produced their
handwriting, including all writers under 18. Handwriting
submitted in response to requests through social media was
subsequently reviewed by the authors. For PromitoLipi1.1,
writers were asked to produce either a coherent paragraph
on a topic of their choice or an incoherent sequence of words.
No particular handwriting style was prescribed. Writers also
produced separate pages of random words for PromitoLipi2.1
using regular stationery.

The handwriting was captured using scanners and smartphone cameras. Named smartphones include OnePlus 5T, OnePlus 7, Google Pixel 6, Poco X3 Pro, and Nokia 8v. Image dimensions and acquisition conditions vary. Thus, the intended variability includes writers' ages and handwriting styles, character size and spacing, paper and writing medium, capture device, illumination, and retained background content.

\section{Preprocessing and annotation}
\subsection{Image preparation}

Images in PromitoLipi1.1 were cropped without denoising.
Ink blots, smudges, uneven stroke thickness, shadows, and background, including parts of a tabletop or
floor, were preserved as captured within the retained
regions. Crossed-out and overwritten text was retained.

Images from all PromitoLipi2.1 sources underwent the same
preparation procedure: segmentation using the method and
implementation of \citet{satter2023segmentation},
followed by manual crop refinement to remove content
outside the target text. The resulting crops were
binarized using Otsu's method, which automatically selects
the threshold. Foreground text was represented by black
pixels (0) and the background by white pixels (255).
No additional stroke smoothing, thinning, or normalization
was applied before annotation.

For the portion of PromitoLipi2.1 derived from
PromitoLipi1.1, the original color pages remain publicly
available in PromitoLipi1.1. Color crops can be generated
by applying the detected word boundaries to these original
color images and saving the crops without binarization.

\par
\Needspace{10\baselineskip}
\begin{wrapfigure}{r}{0.60\textwidth}
\vspace{-5pt}
\centering
\begin{minipage}[t]{.43\linewidth}
\centering\vspace{0pt}
\includegraphics[width=.65\linewidth,height=3.4cm,keepaspectratio]{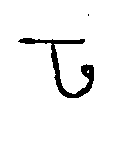}
\par\smallskip
{\small (a) 7435.bmp; single consonant\par}
\bigskip
\includegraphics[width=\linewidth,height=4.5cm,keepaspectratio]{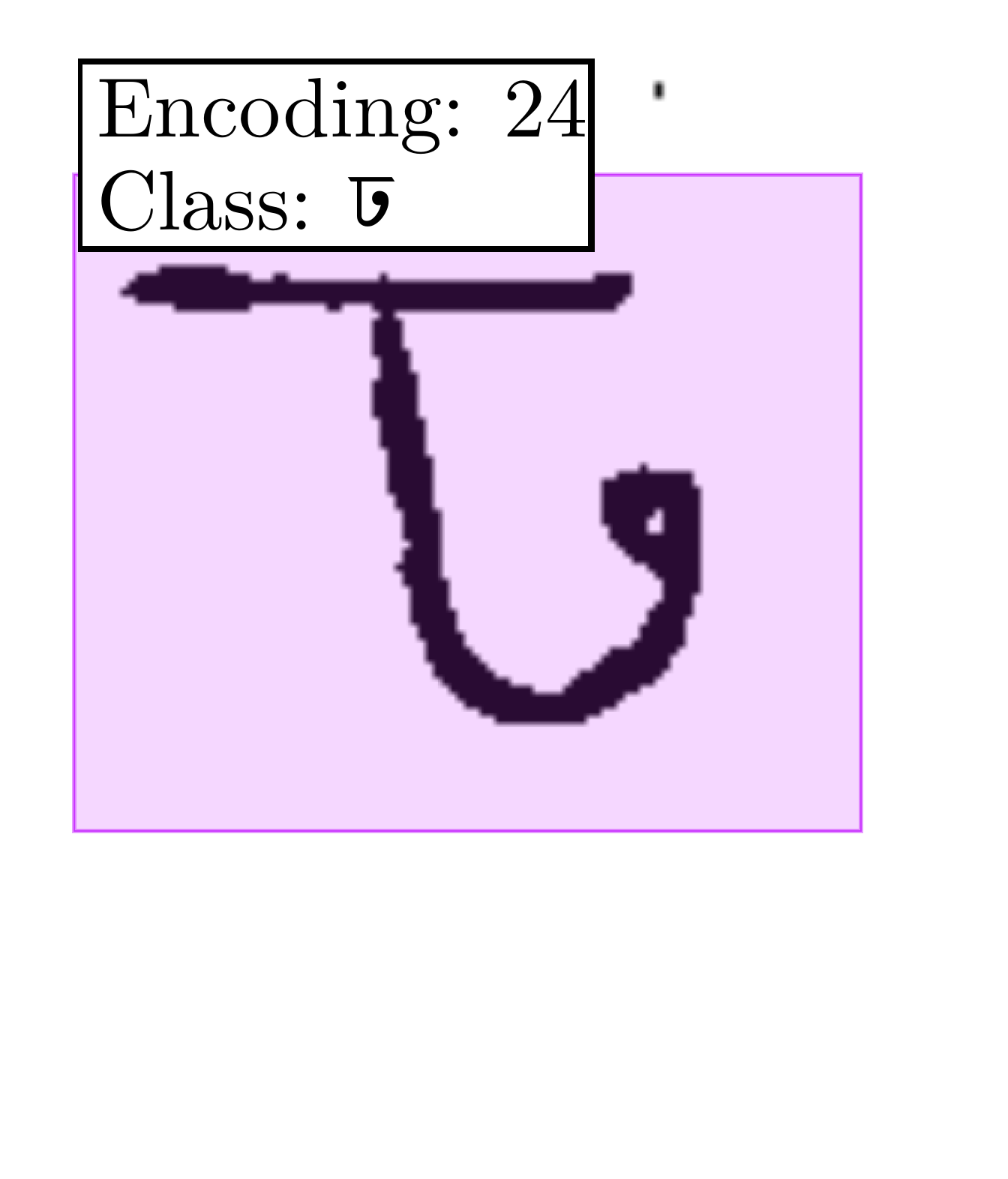}
\par\smallskip
{\small (b) Class 24 and its bounding box\par}
\end{minipage}\hfill
\begin{minipage}[t]{.56\linewidth}
\vspace{0pt}
\begin{lstlisting}[
  language=XML,
  basicstyle=\ttfamily\footnotesize,
  breaklines=true,
  columns=fullflexible,
  aboveskip=0pt
]
<annotation>
  <folder>my-project-name</folder>
  <filename>7435.bmp</filename>
  <path>/my-project-name/7435.bmp</path>
  <source>
    <database>Unspecified</database>
  </source>
  <size>
    <width>135</width>
    <height>159</height>
    <depth>3</depth>
  </size>
  <object>
    <name>24</name>
    <pose>Unspecified</pose>
    <truncated>0</truncated>
    <difficult>0</difficult>
    <bndbox>
      <xmin>12</xmin>
      <ymin>26</ymin>
      <xmax>115</xmax>
      <ymax>112</ymax>
    </bndbox>
  </object>
</annotation>
\end{lstlisting}
\centering
{\small (c) 7435.xml\par}
\end{minipage}
\caption{A PromitoLipi2.1 image, its annotation overlay, and the matching XML record. Class 24 denotes \bn{ঢ}.}
\vspace{-8pt}
\label{fig:xml}
\end{wrapfigure}

\subsection{Classes and annotation format}
All bounding boxes and class IDs in PromitoLipi2.1 were assigned
manually using Make Sense~\cite{makesense}. Each annotated image
was then reviewed by volunteers and authors, who corrected
missing boxes, incorrect class IDs, and poorly fitted boxes.
No formal inter-annotator agreement score was computed.

The XML annotations use 88 class IDs ranging from 1 to 92,
with IDs 65, 80, 86, and 90 unused. Users should follow the
ID-to-class mapping in Appendix~\ref{app:classes} rather
than assume that the classes are numbered consecutively. The scheme represents \bn{আ} through \bn{অ} and \bn{া}, and \bn{ো} through \bn{ে} and \bn{া}~\cite{hasan2021thesis,satter2023recognition}. It is a visual-component scheme rather than a one-to-one Unicode transcription. Converting detections into text requires an ordering and composition rule.

Each VOC-style XML file records the image name, dimensions, and annotated objects. An object contains its class ID in \texttt{name} and bounding-box coordinates \texttt{xmin}, \texttt{ymin}, \texttt{xmax}, and \texttt{ymax}. Figure~\ref{fig:xml} shows an example.

\section{Reported results and code availability}
\label{sec:results}

\citet{satter2023segmentation} reported a word-segmentation
F1-score of 91.20\% on an earlier version of PromitoLipi1.1
containing 80 pages and 7,374 words. On an earlier version
of PromitoLipi2.1 containing 9,841 images, the ensemble
detector of \citet{satter2023recognition} achieved a
character-level macro F1-score of 92.61\% and recognized
72.52\% of the 2,140 test words exactly.

The earlier versions of PromitoLipi1.1 and PromitoLipi2.1
used in the cited studies were further cleaned before
public release, leaving 7,231 words in PromitoLipi1.1 and
9,830 images in PromitoLipi2.1. The published scores have not been recomputed on this release and should be read as indicative rather than as exact benchmarks.

The undergraduate thesis and links to the related publications are available in the \href{https://github.com/FaiasPromit/Optical-Character-Recognition-From-Handwritten-Bangla-Texts}{thesis repository}. Code for \citet{satter2023segmentation} is available in the \href{https://github.com/FaiasPromit/Color-Independent-Word-Segmentation-From-Transcribed-Bangla-Passages}{word segmentation repository}, and code for \citet{satter2023recognition} is available in the \href{https://github.com/FaiasPromit/Open-Vocabulary-Word-Recognition-From-Transcribed-Bangla-Texts}{word recognition repository}.

\section{Limitations and reuse}
PromitoLipi includes varied handwriting and acquisition
conditions, but its scale is limited, particularly in
PromitoLipi1.1, and most writers in the convenience sample
were aged 20--30. Class frequencies are unequal, only
selected conjuncts are included, and artifacts such as
crossed-out and overwritten words were not counted across
the full dataset. 

\section{Ethics and declarations}
Before collecting any handwriting, the first author explained
the purpose of the work---an undergraduate thesis on Bangla
handwriting recognition---and the plans to publish the research
and release the handwriting as a public dataset. Each writer
then gave consent, either in person or through social-media
messages. For writers under 18, permission was obtained
from a parent. The study was not reviewed by an institutional ethics committee.

\paragraph{Author contributions.}
\textbf{Faias Satter:}
Conceptualization, methodology, investigation, formal analysis,
writing---original draft, writing---review and editing.
\textbf{Noor Masrur:}
Investigation, formal analysis,
writing---original draft, writing---review and editing.
\textbf{Sk. Md. Masudul Ahsan:}
Supervision, conceptualization, writing---review and editing.

\paragraph{Funding and acknowledgements.}
This research did not receive a specific grant from funding
agencies in the public, commercial, or not-for-profit sectors.
The authors thank the handwriting contributors and the
volunteers who helped prepare and check the annotations,
including Sajia Abedin from the Department of Pharmaceutical
Sciences at NSU, and Mishu Baidya, Md Kaisarul Islam,
Md. Abu Loman Hossain Shuvo, and other volunteers from the
Department of Computer Science and Engineering at KUET.
They also acknowledge the CMATER research group for the
reused source resources.

\paragraph{Competing interests.}
The authors declare no competing interests.

\FloatBarrier
\bibliographystyle{unsrtnat}
\begingroup\small\raggedright
\bibliography{references}
\endgroup

\clearpage
\appendix
\section{PromitoLipi1.1: paragraph inventory}
\label{app:paragraphs}

PromitoLipi1.1 contains 80 single-page handwritten paragraphs
with 7,231 words. Content is labeled coherent (C) when
sentences are meaningfully related, or incoherent (I)
when it consists of unrelated words. These labels describe
textual coherence and do not indicate image quality.

\begin{table}[htbp]
\centering\small
\caption{Complete inventory of PromitoLipi1.1, showing
each paragraph image's filename, word count, and content
type (C: coherent; I: incoherent).}
\label{tab:paragraphs}
\setlength{\tabcolsep}{5pt}
\begin{tabular}{@{}lrc@{\hspace{9pt}}lrc@{\hspace{9pt}}lrc@{\hspace{9pt}}lrc@{}}
\toprule
File & Words & Type & File & Words & Type & File & Words & Type & File & Words & Type\\\midrule
B001 & 67 & C & B021 & 163 & C & B041 & 115 & C & B061 & 89 & C \\
B002 & 115 & C & B022 & 55 & C & B042 & 54 & C & B062 & 156 & C \\
B003 & 131 & C & B023 & 103 & C & B043 & 118 & C & B063 & 129 & I \\
B004 & 103 & C & B024 & 102 & C & B044 & 72 & C & B064 & 129 & I \\
B005 & 96 & C & B025 & 104 & C & B045 & 37 & C & B065 & 54 & C \\
B006 & 96 & C & B026 & 98 & C & B046 & 44 & C & B066 & 55 & C \\
B007 & 182 & C & B027 & 101 & C & B047 & 43 & C & B067 & 69 & I \\
B008 & 45 & C & B028 & 91 & C & B048 & 48 & C & B068 & 110 & I \\
B009 & 97 & C & B029 & 91 & C & B049 & 135 & C & B069 & 66 & C \\
B010 & 112 & C & B030 & 109 & C & B050 & 95 & C & B070 & 76 & C \\
B011 & 51 & C & B031 & 113 & C & B051 & 93 & C & B071 & 86 & C \\
B012 & 89 & C & B032 & 113 & C & B052 & 139 & C & B072 & 111 & C \\
B013 & 81 & C & B033 & 108 & C & B053 & 105 & C & B073 & 83 & C \\
B014 & 107 & C & B034 & 95 & C & B054 & 17 & C & B074 & 86 & C \\
B015 & 103 & C & B035 & 95 & C & B055 & 19 & C & B075 & 80 & C \\
B016 & 51 & C & B036 & 97 & C & B056 & 14 & C & B076 & 84 & C \\
B017 & 112 & C & B037 & 110 & C & B057 & 84 & C & B077 & 93 & C \\
B018 & 56 & C & B038 & 97 & C & B058 & 82 & C & B078 & 123 & C \\
B019 & 103 & C & B039 & 100 & C & B059 & 96 & C & B079 & 70 & I \\
B020 & 58 & C & B040 & 72 & C & B060 & 71 & C & B080 & 129 & I \\
\bottomrule
\end{tabular}
\par\smallskip\footnotesize All filenames have the extension \texttt{.bmp}. Total: 80 images and 7,231 words.
\end{table}

\clearpage
\section{PromitoLipi2.1: visual-class inventory}
\label{app:classes}

PromitoLipi2.1 contains 9,830 binarized handwriting images
with 24,050 annotated bounding boxes across 88 visual
classes. Class IDs match those used in the XML annotation files. Each occurrence count gives the number of bounding boxes assigned to that class; an image may contain multiple instances.
IPA (International Phonetic Alphabet) glosses illustrate
pronunciation and are not recognition targets.

\begin{table}[htbp]
\centering\small
\caption{Visual-class inventory of PromitoLipi2.1,
showing each class ID, Bangla glyph, IPA gloss, and
number of annotated bounding boxes.}
\label{tab:classes}
\setlength{\tabcolsep}{3pt}
\renewcommand{\arraystretch}{1.10}
\begin{tabular}{@{}rlcr@{\hspace{7pt}}rlcr@{\hspace{7pt}}rlcr@{\hspace{7pt}}rlcr@{}}
\toprule
ID & Glyph & IPA & Count & ID & Glyph & IPA & Count & ID & Glyph & IPA & Count & ID & Glyph & IPA & Count\\\midrule
1 & \bn{অ} & \ipa{/ɔ/} & 334 & 23 & \bn{ড} & \ipa{/d/} & 165 & 45 & \bn{য়} & \ipa{/ʲ/} & 261 & 68 & \bn{ক্ষ} & \ipa{/kkʰo/} & 161\\
2 & \bn{ই} & \ipa{/i/} & 207 & 24 & \bn{ঢ} & \ipa{/dʱɔ/} & 146 & 46 & \bn{ৎ} & \ipa{/t̪/} & 121 & 69 & \bn{ন্ত} & \ipa{/nt̪o/} & 180\\
3 & \bn{ঈ} & \ipa{/i/} & 107 & 25 & \bn{ণ} & \ipa{/n/} & 199 & 47 & \bn{ঃ} & \ipa{/h/} & 104 & 70 & \bn{ত্র} & \ipa{/t̪ro/} & 135\\
4 & \bn{উ} & \ipa{/u/} & 182 & 26 & \bn{ত} & \ipa{/t̪o/} & 549 & 48 & \bn{ং} & \ipa{/ŋ/} & 142 & 71 & \bn{ঙ্গ} & \ipa{/ŋgo/} & 126\\
5 & \bn{ঊ} & \ipa{/u/} & 120 & 27 & \bn{থ} & \ipa{/t̪ʰo/} & 149 & 49 & \bn{ঁ} & \ipa{/ ̃/} & 128 & 72 & \bn{স্থ} & \ipa{/st̪ʰɔ/} & 181\\
6 & \bn{ঋ} & \ipa{/ri/} & 173 & 28 & \bn{দ} & \ipa{/d̪/} & 427 & 50 & \bn{০} & \ipa{/0/} & 323 & 73 & \bn{স্ব} & \ipa{/ʃɔ/} & 133\\
7 & \bn{এ} & \ipa{/e/} & 217 & 29 & \bn{ধ} & \ipa{/d̪ʱɔ/} & 188 & 51 & \bn{১} & \ipa{/1/} & 338 & 74 & \bn{ক্ত} & \ipa{/kt̪o/} & 125\\
8 & \bn{ঐ} & \ipa{/o͡i̯/} & 119 & 30 & \bn{ন} & \ipa{/n/} & 731 & 52 & \bn{২} & \ipa{/2/} & 282 & 75 & \bn{স্ত} & \ipa{/st̪o/} & 139\\
9 & \bn{ও} & \ipa{/o/} & 132 & 31 & \bn{প} & \ipa{/p/} & 394 & 53 & \bn{৩} & \ipa{/3/} & 322 & 76 & \bn{ন্দ} & \ipa{/nd̪o/} & 175\\
10 & \bn{ঔ} & \ipa{/o͡u̯/} & 121 & 32 & \bn{ফ} & \ipa{/pʰo/} & 139 & 54 & \bn{৪} & \ipa{/4/} & 319 & 77 & \bn{চ্ছ} & \ipa{/ccʰo/} & 160\\
11 & \bn{ক} & \ipa{/k/} & 657 & 33 & \bn{ব} & \ipa{/bo/} & 640 & 55 & \bn{৫} & \ipa{/5/} & 322 & 78 & \bn{দ্ধ} & \ipa{/d̪d̪ʱo/} & 137\\
12 & \bn{খ} & \ipa{/kʰ/} & 223 & 34 & \bn{ভ} & \ipa{/bʱɔ/} & 124 & 56 & \bn{৬} & \ipa{/6/} & 273 & 79 & \bn{ন্ত্র} & \ipa{/nt̪ro/} & 109\\
13 & \bn{গ} & \ipa{/go/} & 293 & 35 & \bn{ম} & \ipa{/m/} & 565 & 57 & \bn{৭} & \ipa{/7/} & 252 & 81 & \bn{ত্ত} & \ipa{/t̪t̪o/} & 134\\
14 & \bn{ঘ} & \ipa{/gʱɔ/} & 152 & 36 & \bn{য} & \ipa{/ɟo/} & 144 & 58 & \bn{৮} & \ipa{/8/} & 322 & 82 & \bn{ষ্ট} & \ipa{/ʃto/} & 188\\
15 & \bn{ঙ} & \ipa{/ŋ/} & 106 & 37 & \bn{র} & \ipa{/ro/} & 768 & 59 & \bn{৯} & \ipa{/9/} & 303 & 83 & \bn{ন্ন} & \ipa{/nno/} & 195\\
16 & \bn{চ} & \ipa{/c/} & 218 & 38 & \bn{ল} & \ipa{/lɔ/} & 407 & 60 & \bn{া} & \ipa{/a/} & 2040 & 84 & \bn{ল্প} & \ipa{/lpo/} & 135\\
17 & \bn{ছ} & \ipa{/cʰɔ/} & 149 & 39 & \bn{শ} & \ipa{/ʃɔ/} & 293 & 61 & \bn{ি} & \ipa{/i/} & 861 & 85 & \bn{ম্প} & \ipa{/mp/} & 134\\
18 & \bn{জ} & \ipa{/ɟ/} & 231 & 40 & \bn{ষ} & \ipa{/ʃɔ/} & 181 & 62 & \bn{ী} & \ipa{/i/} & 215 & 87 & \bn{ূ} & \ipa{/u/} & 104\\
19 & \bn{ঝ} & \ipa{/ɟʱɔ/} & 112 & 41 & \bn{স} & \ipa{/ʃɔ/} & 393 & 63 & \bn{ে} & \ipa{/e/} & 1092 & 88 & \bn{ৃ} & \ipa{/ri/} & 99\\
20 & \bn{ঞ} & \ipa{/n/} & 122 & 42 & \bn{হ} & \ipa{/ho/} & 360 & 64 & \bn{ু} & \ipa{/u/} & 386 & 89 & \bn{ৈ} & \ipa{/o͡i̯/} & 137\\
21 & \bn{ট} & \ipa{/t/} & 241 & 43 & \bn{ড়} & \ipa{/ɽ/} & 183 & 66 & \bn{্র} & \ipa{/r/} & 668 & 91 & \bn{ৌ} & \ipa{/o͡u̯/} & 151\\
22 & \bn{ঠ} & \ipa{/tʰ/} & 134 & 44 & \bn{ঢ়} & \ipa{/ɽʱo/} & 107 & 67 & \bn{্য} & \ipa{/ː/} & 234 & 92 & \bn{।} & \ipa{/\bn{।}/} & 102\\
\bottomrule
\end{tabular}
\par\smallskip\footnotesize Total: 24,050 instances. Range: 99--2,040 instances per class. IDs 65, 80, 86, and 90 are not entries in this inventory.
\end{table}

\end{document}